\documentclass[conference]{IEEEtran}
\IEEEoverridecommandlockouts
\usepackage{cite}
\usepackage{amsmath,amssymb,amsfonts}
\usepackage{algorithm}
\usepackage{algorithmic}
\usepackage{subcaption}
\usepackage{multirow}
\usepackage{colortbl}

\usepackage{graphicx} 
\usepackage{booktabs} 
\usepackage{textcomp}
\usepackage{xcolor}
\usepackage{hyperref}
\usepackage{xcolor,tcolorbox}
\usepackage{amsmath}
\usepackage{siunitx}
\usepackage{threeparttable} 

\def\BibTeX{{\rm B\kern-.05em{\sc i\kern-.025em b}\kern-.08em
    T\kern-.1667em\lower.7ex\hbox{E}\kern-.125emX}}

\begin{document}

\title{Cognitive-Mental-LLM: Evaluating Reasoning in Large Language Models for Mental Health Prediction via Online Text}

\author{\IEEEauthorblockN{1\textsuperscript{st} Avinash Patil}
\IEEEauthorblockA{\textit{Ira A. Fulton Schools of Engineering} \\
\textit{Arizona State University}\\
Tempe, USA \\
avinashpatil@ieee.org}
\and
\IEEEauthorblockN{2\textsuperscript{nd} Amardeep Kour Gedhu}
\IEEEauthorblockA{\textit{Department of Psychology} \\
\textit{Santa Clara University}\\
Santa Clara, USA \\
agedhu@scu.edu}
}
\maketitle

\begin{abstract}
Large Language Models (LLMs) have demonstrated potential in predicting mental health outcomes from online text, yet traditional classification methods often lack interpretability and robustness. This study evaluates structured reasoning techniques—Chain-of-Thought (CoT), Self-Consistency (SC-CoT), and Tree-of-Thought (ToT)—to improve classification accuracy across multiple mental health datasets sourced from Reddit. We analyze reasoning-driven prompting strategies, including Zero-shot CoT and Few-shot CoT, using key performance metrics such as Balanced Accuracy, F1 score, and Sensitivity/Specificity. Our findings indicate that reasoning-enhanced techniques improve classification performance over direct prediction, particularly in complex cases. Compared to baselines such as Zero Shot non-CoT Prompting, and fine-tuned pre-trained transformers such as BERT and Mental-RoBerta, and fine-tuned Open Source LLMs such as Mental Alpaca and Mental-Flan-T5, reasoning-driven LLMs yield notable gains on datasets like Dreaddit (+0.52\% over M-LLM, +0.82\% over BERT) and SDCNL (+4.67\% over M-LLM, +2.17\% over BERT). However, performance declines in Depression Severity, and CSSRS predictions suggest dataset-specific limitations, likely due to our using a more extensive test set. Among prompting strategies, Few-shot CoT consistently outperforms others, reinforcing the effectiveness of reasoning-driven LLMs. Nonetheless, dataset variability highlights challenges in model reliability and interpretability. This study provides a comprehensive benchmark of reasoning-based LLM techniques for mental health text classification. It offers insights into their potential for scalable clinical applications while identifying key challenges for future improvements. Code, prompts and llm reasoning for classification are available at \url{https://github.com/av9ash/cognitive_mental_llm}

\end{abstract}

\begin{IEEEkeywords}
Large Language Models, Reasoning, Mental Health Prediction, Chain-of-Thought, Self-consistency, Tree-of-thought, Few-shot learning, Natural Language Processing, Online Text Analysis
\end{IEEEkeywords}

\section{Introduction}

Mental health disorders, such as depression, anxiety, and suicidal ideation, represent a growing global concern, with millions of individuals affected annually \cite{world2022mental,perou2013mental}. Online platforms, mainly social media and mental health forums, have become vital spaces for individuals to express their emotions and seek support \cite{de2014mental}. This has led to a surge in interest in AI-driven methods for analyzing and classifying mental health-related text. However, accurately interpreting such text remains a significant challenge due to the complexity and variability of natural language in mental health discourse \cite{calvo2017natural}.

Traditional NLP-based classifiers, such as BERT \cite{devlin2019bert} and RoBERTa \cite{liu2019roberta}, have demonstrated strong performance in general text classification but often struggle with mental health data due to subtle linguistic cues, contextual ambiguity, and the need for structured reasoning \cite{jadon2025enhancing}. Prior studies \cite{xu2024mental, turcan2019dreaddit, gaur2019knowledge} have explored deep learning and transformer-based approaches, yet these models exhibit limitations in interpretability and robustness when applied to real-world mental health assessments.

To address these challenges, we investigate the role of structured reasoning in mental health classification using OpenAI’s \textit{o3-mini}, a small reasoning-focused large language model (LLM). Our study evaluates four structured reasoning prompting strategies: \textbf{Chain-of-Thought (CoT) \cite{wei2022chain}, Self-Consistency CoT (SC-CoT) \cite{wang2022self}, Few-Shot CoT \cite{brown2020language}, and Tree-of-Thought (ToT) \cite{yao2023tree}}. These approaches encourage the model to generate step-by-step reasoning before classification, improving robustness and interpretability.

\textbf{Our main contributions are as follows:}
\begin{itemize}
    \item We apply structured reasoning techniques to mental health text classification and evaluate their effectiveness across five benchmark datasets: \textbf{Dreaddit} \cite{turcan2019dreaddit}, \textbf{CSSRS} \cite{gaur2019knowledge}, \textbf{SDCNL} \cite{haque2021deep}, \textbf{DepSeverity} \cite{naseem2022early}, and \textbf{RedSam} \cite{sampath2022data}.
    \item We compare reasoning-based prompting strategies against zero-shot classification and prior state-of-the-art transformer models (\textbf{BERT, RoBERTa, Alpaca, FLAN-T5}) \cite{xu2024mental}.
    \item We conduct a detailed analysis of classification performance, showing that Few-Shot CoT improves multi-class classification tasks while CoT and SC-CoT enhance binary classification robustness.
\end{itemize}

Our results demonstrate that structured reasoning strategies improve classification accuracy, particularly in datasets with nuanced language and multi-class labels. Notably, Few-Shot CoT performs superior in CSSRS and DepSeverity, while CoT and SC-CoT enhance classification for Dreaddit and SDCNL.

\textbf{Paper Organization:} The remainder of this paper is structured as follows: Section~\ref{sec:related} reviews related work. Section~\ref{sec:methodology} describes our methodology, including datasets and prompting techniques. Section~\ref{sec:results} presents experimental results, and Section~\ref{sec:conclusion} concludes with key findings and future directions.

\section{Related Work}
\label{sec:related}
The application of Large Language Models (LLMs) in mental health prediction has garnered significant attention in recent years. Traditional machine learning approaches, such as Support Vector Machines (SVMs) and logistic regression, have been employed for mental health text classification \cite{wongkoblap2017researching}. However, these models often require extensive feature engineering and struggle with capturing contextual nuances in text data.

With the advent of transformer-based architectures, LLMs have demonstrated improved capabilities in understanding and generating human-like text. Xu et al. \cite{xu2024mental} introduced \textit{Mental-LLM}, leveraging LLMs for mental health prediction via online text data. Their study highlighted that LLMs outperform traditional methods in predictive accuracy. However, these models primarily rely on direct classification and often lack interpretability, which is crucial in mental health assessments.

Several prompting techniques have been proposed to enhance reasoning capabilities and interpretability. Wei et al. \cite{wei2022chain} introduced Chain-of-Thought (CoT) prompting, enabling LLMs to generate intermediate reasoning steps, improving performance on complex tasks. Building upon this, Wang et al. \cite{wang2022self} proposed Self-Consistency (SoTC), which involves generating multiple reasoning paths and selecting the most consistent answer, enhancing reliability. Yao et al. \cite{yao2023tree} extended these concepts with Tree-of-Thought (ToT) prompting, introducing a structured, hierarchical reasoning process for deliberate problem-solving.

Recent research has explored the ability of large language models (LLMs) to evaluate responses to suicidal ideation \cite{mcbain2025competency}. In an observational, cross-sectional study, three widely used LLMs—ChatGPT-4o, Claude 3.5 Sonnet, and Gemini 1.5 Pro—were assessed on their capacity to rate clinician responses from the revised Suicidal Ideation Response Inventory (SIRI-2). The study compared LLM-generated ratings to expert suicidologists' ratings using linear regression analyses and z-score outlier detection.

Recent advancements have also explored the integration of reasoning and acting within LLMs. Yao et al. \cite{yao2023react} proposed the ReAct pattern, synergizing reasoning and acting in language models to improve task performance. This approach allows LLMs to reason through problems and take actions based on their reasoning, leading to more effective problem-solving strategies.

Despite these advancements, applying reasoning techniques in LLMs to mental health text classification remains underexplored. Patil \cite{patil2025advancing} provides a comprehensive overview of promising methods and approaches in advancing reasoning in LLMs, emphasizing their potential to enhance model interpretability and decision-making. However, there is a lack of studies systematically evaluating the impact of these reasoning techniques on mental health prediction tasks.

In this study, we aim to bridge the gap in mental health text classification by systematically evaluating structured reasoning techniques—Chain-of-Thought (CoT), Self-Consistency (SC-CoT), Few-Shot Learning with CoT (FS-CoT), and Tree-of-Thought (ToT). We benchmark these reasoning-driven approaches against established models, including BERT, RoBERTa, and the best-performing supervised fine-tuned large language models from \cite{xu2024mental}. Our comprehensive analysis spans multiple datasets, providing deeper insights into the effectiveness of reasoning techniques in enhancing LLM-based mental health assessments.

\section{Methodology}x
\label{sec:methodology}
This study evaluates the effectiveness of structured reasoning techniques for mental health text classification using OpenAI’s \textit{o3-mini}\cite{o3-mini} model, a compact, reasoning-oriented language model known for its strong performance in scientific, mathematical, and programming tasks. We examine four prompting strategies—Chain-of-Thought (CoT), Tree-of-Thought (ToT), Few-shot CoT, and Self-Consistency (SC-CoT)—each enabling the model to generate interpretable reasoning prior to classification. This approach enhances both robustness and accuracy in mental health assessments. For comparison, we include results from fine-tuned models documented in a previous\cite{xu2024mental} study; however, no fine-tuning was conducted here, as our primary focus lies in evaluating the impact of different reasoning strategies.

\subsection{Data Collection and Preprocessing}
\begin{figure*}[htbp]
    \centering
    \caption{Class Distributions Across Different Mental Health Datasets}
    \begin{subfigure}[b]{0.32\textwidth}
        \includegraphics[width=\textwidth]{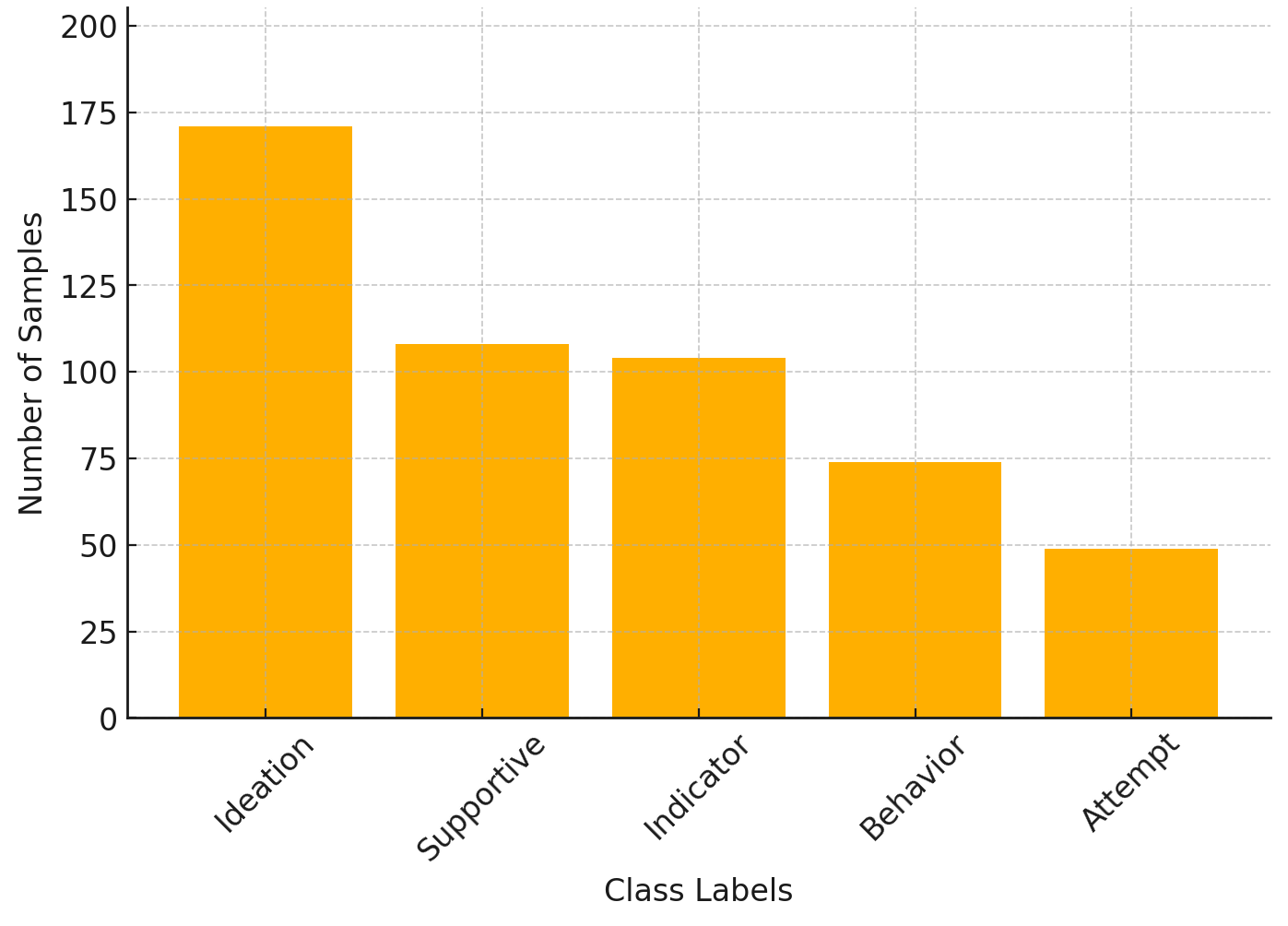}
        \caption{CSSRS}
        \label{fig:cssrs}
    \end{subfigure}
    \hfill
    \begin{subfigure}[b]{0.32\textwidth}
        \includegraphics[width=\textwidth]{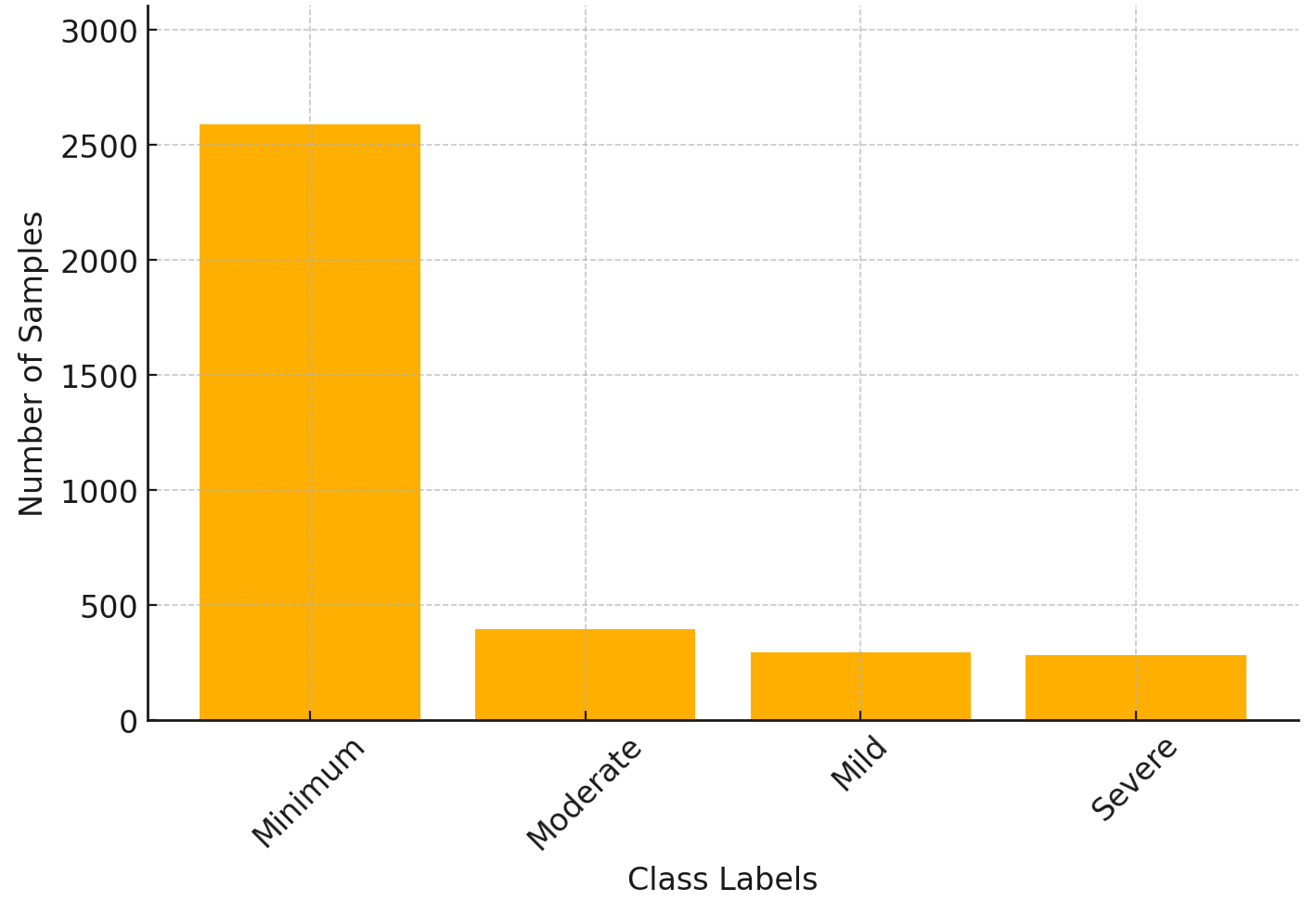}
        \caption{DepSeverity}
        \label{fig:depseverity}
    \end{subfigure}
    \hfill
    \begin{subfigure}[b]{0.32\textwidth}
        \includegraphics[width=\textwidth]{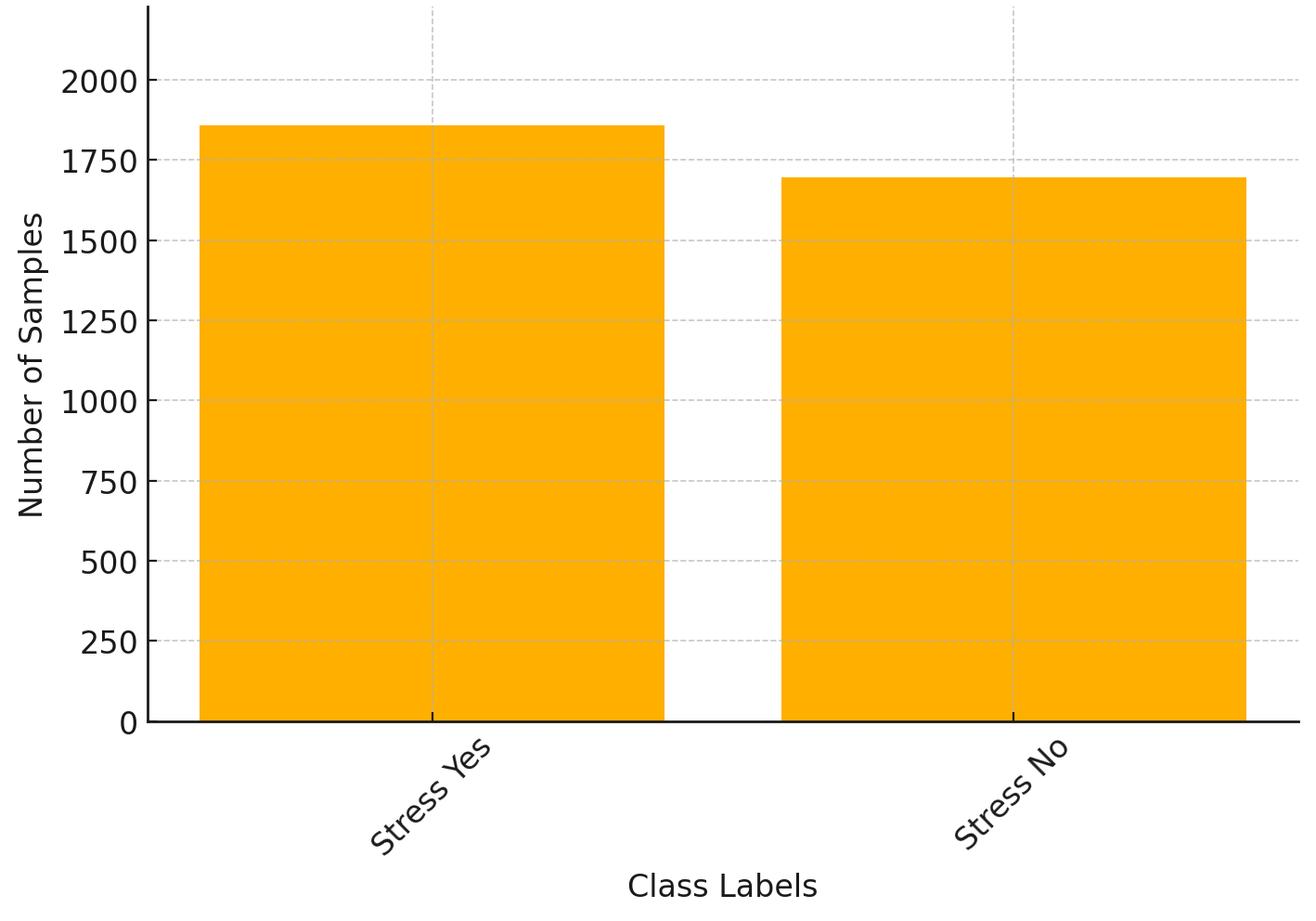}
        \caption{Dreaddit}
        \label{fig:dreaddit}
    \end{subfigure}
    \hfill
    \begin{subfigure}[b]{0.32\textwidth}
        \includegraphics[width=\textwidth]{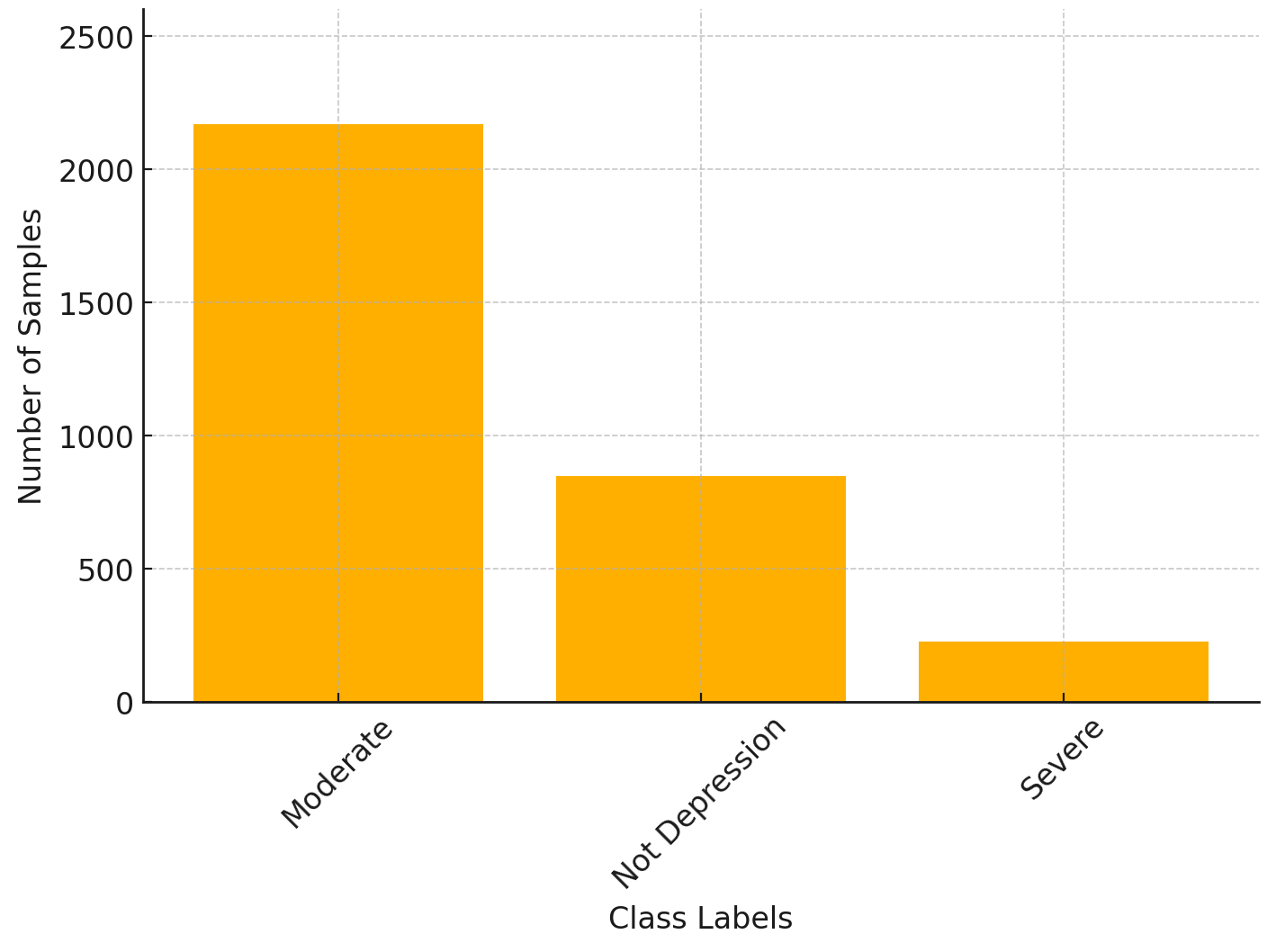}
        \caption{RedSam}
        \label{fig:redsam}
    \end{subfigure}
    \hfill
    \begin{subfigure}[b]{0.32\textwidth}
        \includegraphics[width=\textwidth]{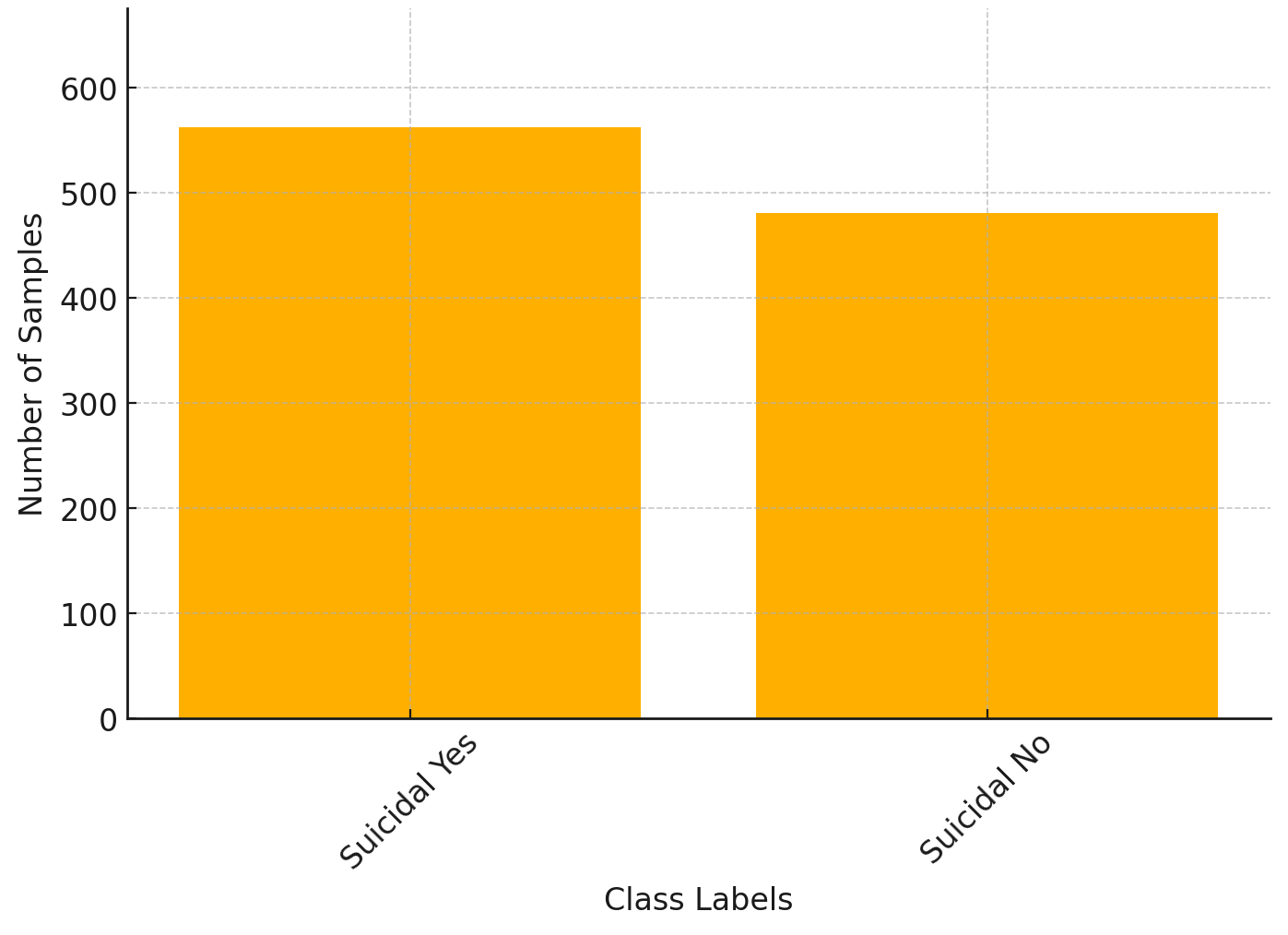}
        \caption{SDCNL}
        \label{fig:sdcnl}
    \end{subfigure}
    \label{fig:dataset_distributions}
\end{figure*}

We use five benchmark datasets from previous studies, each containing online text relevant to mental health assessments. While prior studies partitioned these datasets into train-test splits, our study focuses on zero-shot classification. Therefore, we utilize the entire dataset for testing except for RedSam, ensuring a more comprehensive evaluation. Table \ref{tab:dataset_summary} provides an overview of these datasets. Figure \ref{fig:dataset_distributions} provides an overview of these distributions.

\begin{itemize}
    \item \textbf{Dreaddit} \cite{turcan2019dreaddit} – This dataset consists of posts from ten subreddits across five domains: abuse, social, anxiety, PTSD, and financial. It includes 2,929 user posts, with multiple human annotators assessing whether specific sentence segments indicate user stress. The final labels were obtained by aggregating these annotations.  

    \item \textbf{CSSRS} \cite{gaur2019knowledge} – Containing posts from 15 mental health-related subreddits, this dataset includes data from 2,181 users collected between 2005 and 2016. Four psychiatrists manually annotated posts from 500 users based on the Columbia Suicide Severity Rating Scale (C-SSRS) \cite{posner2008columbia}, categorizing suicide risk into five levels: supportive, indicator, ideation, behavior, and attempt. However, the user-generated text in this dataset can be highly variable and may introduce significant noise.

    \item \textbf{SDCNL} \cite{haque2021deep} – This dataset comprises posts from r/SuicideWatch and r/Depression, contributed by 1,723 users. Each post was manually annotated to indicate the presence of suicidal thoughts. To ensure cost-effective processing and focus on concise textual data, we limit our selection to posts with fewer than 128 words, resulting in a refined dataset of 1,044 posts.

    \item \textbf{DepSeverity} \cite{naseem2022early} – Two human annotators classified posts from the Dreaddit dataset \cite{turcan2019dreaddit} into four levels of depression—minimal, mild, moderate, and severe—based on the DSM-5 \cite{regier2013dsm} guidelines.  

    \item \textbf{RedSam} \cite{sampath2022data} – This dataset contains posts from five mental health-related subreddits: Mental Health, Depression, Loneliness, Stress, and Anxiety. Depression labels were derived by aggregating annotations from two domain experts. Due to this dataset's large number of posts, we use only the test set for this model as other studies.
\end{itemize}

\begin{table*}[t]
\centering
\caption{Summary of Mental Health Datasets: Size, Class Distribution, and Text Length}
\begin{tabular}{l r r@{ $\pm$ }l r}
\toprule
\textbf{Dataset} & \textbf{Size} & \multicolumn{2}{c}{\textbf{Text Length (Mean $\pm$ Std)}} & \textbf{Class Distribution (Order: Highest to Lowest)} \\
\midrule
\textbf{CSSRS}    & 500    & 1,344 & 1,640 & Ideation: 34.2\%, Supportive: 21.6\%, Indicator: 20.8\%, Behavior: 14.8\%, Attempt: 9.8\% \\
\textbf{DepSeverity} & 3,553 &   86  &   32  & Minimum: 72.8\%, Moderate: 11.1\%, Mild: 8.2\%, Severe: 7.9\% \\
\textbf{Dreaddit}  & 3,553 &   86  &   32  & Stress Yes: 52.3\%, Stress No: 47.7\% \\
\textbf{SDCNL}    & 1,044 &   60  &   33  & Suicidal Yes: 53.9\%, Suicidal No: 46.1\% \\
\textbf{RedSam}   & 3,245 &  166  &  203  & Moderate: 66.8\%, Not Depression: 26.1\%, Severe: 7.0\% \\
\bottomrule
\end{tabular}
\label{tab:dataset_summary}
\end{table*}
\subsection{Model Architecture and Techniques}
We employ \textit{o3-mini}, OpenAI’s first small reasoning LLM, designed to balance speed and accuracy. Unlike conventional transformer-based classifiers, \textit{o3-mini} leverages structured reasoning for classification tasks. While the model inherently performs reasoning, Chain-of-Thought (CoT) prompting further enforces explicit step-by-step reasoning, mitigates shortcut biases, and enhances interpretability in complex cases. The following prompting techniques are applied:

\begin{itemize}
    \item \textbf{Chain-of-Thought (CoT) \cite{wei2022chain}}: CoT prompting enables language models to generate intermediate reasoning steps before arriving at a final classification. By structuring responses step-by-step, CoT enhances interpretability and improves the model’s ability to handle complex decision-making tasks.

    \item \textbf{Self-Consistency CoT (SC-CoT) \cite{wang2022self}}: SC-CoT builds upon the CoT framework by generating multiple independent reasoning paths for the same query. The model then selects the most consistent prediction based on aggregated responses, improving reliability and reducing variance in decision-making.

    \item \textbf{Few-shot CoT \cite{brown2020language}}: This approach integrates few-shot learning with CoT prompting by providing limited annotated examples as in-context demonstrations. Few-shot learning allows the model to generalize across tasks with minimal supervision, while CoT ensures structured reasoning for classification.

    \item \textbf{Tree-of-Thought (ToT) \cite{yao2023tree}}: ToT extends CoT by introducing a hierarchical reasoning structure. Instead of a linear reasoning process, ToT allows the model to explore multiple potential reasoning paths in a tree-like manner, enabling strategic lookahead and backtracking to refine decision-making in complex classification tasks.
\end{itemize}

These techniques enhance model interpretability, reduce hallucination, and improve classification accuracy across mental health datasets.

\subsection{Experimental Setup}
All experiments were conducted using OpenAI's API with \textit{o3-mini}. The experimental setup is as follows:

\begin{itemize}
    \item \textbf{Model}: OpenAI \textit{o3-mini}
    \item \textbf{Temperature}: 0.7 (balancing diversity and consistency)
    \item \textbf{Max Tokens}: Set to allow for detailed reasoning
\end{itemize}

For each dataset, we evaluate the performance across different reasoning techniques.

\subsection{Evaluation Metrics}
We evaluate performance using the following metrics:

\begin{itemize}
    \item \textbf{Accuracy}: Measures the overall correctness of predictions, representing the proportion of correctly classified instances:
    \[
    \frac{\text{TP} + \text{TN}}{\text{TP} + \text{TN} + \text{FP} + \text{FN}}
    \]

    \item \textbf{Balanced Accuracy (Macro)}: Computes the average sensitivity (recall) across all classes, ensuring that class imbalance does not skew the evaluation:
    \[
    \frac{1}{C}\sum_{c=1}^C \frac{\text{TP}_c}{\text{TP}_c + \text{FN}_c}
    \]
    where \(C\) is the number of classes.
    
    \item \textbf{Precision (Macro)}: Measures the fraction of correctly predicted positive instances among all predicted positives, averaged across classes:
    \[
    \frac{1}{C}\sum_{c=1}^C \frac{\text{TP}_c}{\text{TP}_c + \text{FP}_c}
    \]

    \item \textbf{Recall (Macro)}: Also known as sensitivity, recall quantifies the proportion of actual positive instances that are correctly identified, averaged across classes:
    \[
    \frac{1}{C}\sum_{c=1}^C \frac{\text{TP}_c}{\text{TP}_c + \text{FN}_c}
    \]

    \item \textbf{F1-Score (Macro)}: Computes the harmonic mean of precision and recall for each class and then averages them, balancing precision-recall trade-offs:
    \[
    \frac{1}{C}\sum_{c=1}^C 2 \times \frac{\text{Precision}_c \times \text{Recall}_c}{\text{Precision}_c + \text{Recall}_c}
    \]

    \item \textbf{Matthews Correlation Coefficient (MCC)}: Provides a balanced measure for binary classification, even under imbalanced datasets, by considering all confusion matrix elements:
    \[
    \frac{\text{TP} \times \text{TN} - \text{FP} \times \text{FN}}{\sqrt{(\text{TP}+\text{FP})(\text{TP}+\text{FN})(\text{TN}+\text{FP})(\text{TN}+\text{FN})}}
    \]

    \item \textbf{Mean Absolute Error (MAE)}: Measures the average absolute difference between true and predicted values, commonly used for regression tasks:
    \[
    \frac{1}{N}\sum_{i=1}^N |y_i - \hat{y}_i|
    \]
    where \(y_i\) = true value, \(\hat{y}_i\) = predicted value.

    \item \textbf{Quadratic Weighted Kappa (QWK)}: Evaluates the agreement between predicted and true labels, penalizing larger disagreements more heavily:
    \[
    1 - \frac{\sum_{i,j} w_{ij} O_{ij}}{\sum_{i,j} w_{ij} E_{ij}}
    \]
    with \(w_{ij} = (i-j)^2\), \(O_{ij}\) = observed counts, \(E_{ij}\) = expected counts.

    \item \textbf{ROC AUC}: Represents the area under the Receiver Operating Characteristic (ROC) curve, indicating the model's ability to distinguish between classes.

    \item \textbf{PR AUC}: Measures the area under the Precision-Recall curve, particularly useful for evaluating models in imbalanced classification scenarios, where positive class performance is critical.
\end{itemize}

\textit{Notation}: TP = True Positives, TN = True Negatives, FP = False Positives, FN = False Negatives.

\subsection{Reproducibility}
To ensure reproducibility, we take the following measures:
\begin{itemize}
    \item \textbf{Use of public datasets}: All datasets used are publicly available from prior studies.
    \item \textbf{Standardized prompts}: The reasoning strategies are defined using fixed, reproducible prompt templates.
    \item \textbf{Code release}: Code, Prompts and LLM Reasoning for classification are all available on Github.
\end{itemize}
This study facilitates future research in reasoning-driven mental health classification by maintaining transparency in methodology and dataset usage.

\section{Results and Analysis}
\label{sec:results}

This section presents the classification performance of reasoning-based prompting strategies applied to OpenAI's \textit{o3-mini} model across five mental health benchmark datasets. We evaluate the impact of \textbf{Chain-of-Thought (CoT), Self-Consistency CoT (SC-CoT), Few-Shot CoT (FS-CoT), and Tree-of-Thought (ToT)}, comparing them with zero-shot prompting and prior state-of-the-art models, including \textit{BERT}, \textit{RoBERTa}, \textit{Alpaca}, and \textit{FLAN-T5}.

\subsection{Classification Accuracy}
Table~\ref{tab:results} summarizes classification accuracy across datasets. Key observations include:

\begin{itemize}
    \item \textbf{Dreaddit}: CoT (\textbf{0.791}) outperforms zero-shot by \textbf{6.6\%}, though \textit{RoBERTa} achieves the highest accuracy (\textbf{0.831}).
    \item \textbf{DepSeverity}: Few-Shot CoT (\textbf{0.427}) performs best among reasoning-based methods but remains below zero-shot (\textbf{0.656}) and \textit{BERT} (\textbf{0.690}).
    \item \textbf{CSSRS}: Few-Shot CoT (\textbf{0.469}) surpasses both zero-shot (\textbf{0.441}) and \textit{BERT} (\textbf{0.332}).
    \item \textbf{SDCNL}: CoT (\textbf{0.699}) provides a stable improvement over zero-shot (\textbf{0.647}) and \textit{BERT} (\textbf{0.678}).
    \item \textbf{RedSam}: Zero-shot prompting (\textbf{0.511}) outperforms all reasoning-based techniques.
\end{itemize}

Overall, \textbf{Few-Shot CoT benefits multi-class classification} (CSSRS, DepSeverity), while \textbf{CoT and SC-CoT improve binary classification} (Dreaddit, SDCNL).

\begin{figure}[htbp]
    \centering
    \includegraphics[width=0.9\linewidth]{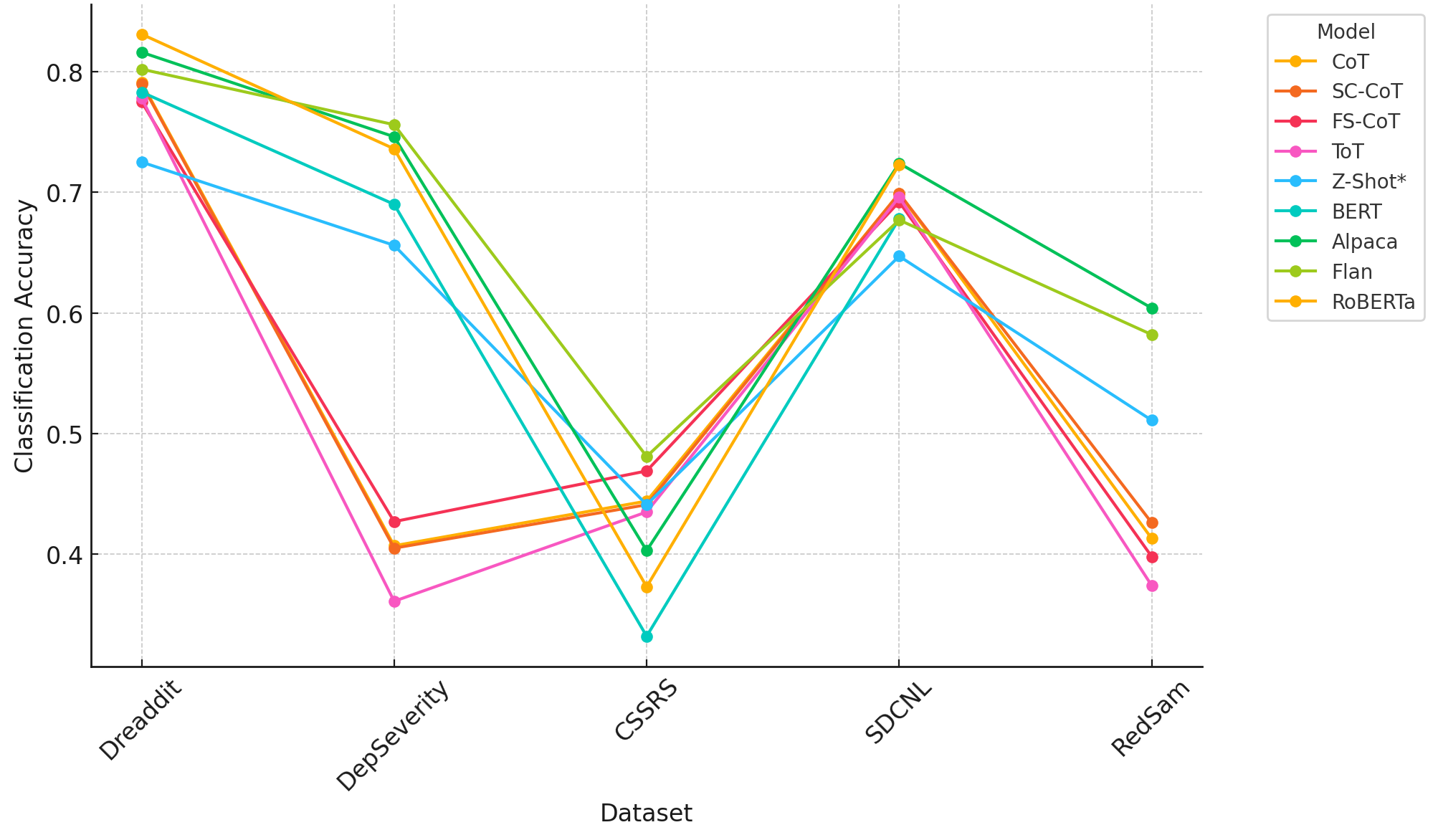}
    \caption{Classification accuracy comparison between reasoning strategies (CoT, SC-CoT, FS-CoT, ToT) and baseline models (BERT, RoBERTa) across five mental health datasets, showing superior performance of Few-Shot CoT on CSSRS and DepSeverity (multi-class tasks).}
    \label{fig:acctrends}
\end{figure}

Figure~\ref{fig:acctrends} illustrates classification accuracy trends across datasets, highlighting the comparative performance of reasoning-based strategies and baseline models.

\begin{table*}[htbp]
    \centering
    \small
    \renewcommand{\arraystretch}{1.2}
    \setlength{\tabcolsep}{4pt}
    \caption{Classification Accuracy and Comparative Gains ($\Delta$) Across Models}
    \label{tab:results}
    \begin{threeparttable}
    \begin{tabular}{@{}l *{11}{c}@{}}
        \toprule
        \textbf{Dataset} 
           & \textbf{CoT} 
           & \textbf{SC-CoT} 
           & \textbf{FS-CoT} 
           & \textbf{ToT}
           & \textbf{Z-Shot}\tnote{*} 
           & \textbf{$\Delta$Z-Shot}
           & \textit{BERT}\tnote{*\dag}
           & \textbf{$\Delta$BERT}
           & \textit{Alpaca}\tnote{*\dag} 
           & \textit{FLAN-T5}\tnote{*\dag} 
           & \textit{RoBERTa}\tnote{*\dag} \\
        \midrule
        Dreaddit
           & 0.791 & 0.790 & 0.775 & 0.778 
           & 0.725 & \textcolor{green}{0.066 ↑} 
           & 0.783 & \textcolor{green}{0.008 ↑} 
           & 0.816 & 0.802 & 0.831 \\
        DepSeverity
           & 0.407 & 0.405 & 0.427 & 0.361 
           & 0.656 & \textcolor{red}{-0.229 ↓} 
           & 0.690 & \textcolor{red}{-0.263 ↓} 
           & 0.746 & 0.756 & 0.736 \\
        CSSRS
           & 0.444 & 0.441 & 0.469 & 0.435 
           & 0.441 & \textcolor{green}{0.028 ↑} 
           & 0.332 & \textcolor{green}{0.137 ↑} 
           & 0.403 & 0.481 & 0.373 \\
        SDCNL
           & 0.699 & 0.699 & 0.692 & 0.696 
           & 0.647 & \textcolor{green}{0.052 ↑} 
           & 0.678 & \textcolor{green}{0.021 ↑} 
           & 0.724 & 0.677 & 0.723 \\
        RedSam
           & 0.413 & 0.426 & 0.398 & 0.374 
           & 0.511 & \textcolor{red}{-0.085 ↓} 
           & -- & -- 
           & 0.604 & 0.582 & -- \\
        \bottomrule
    \end{tabular}
    
    \begin{tablenotes}[flushleft]
        \footnotesize
        \item[*] Results from prior study \cite{xu2024mental}
        \item[\dag] Mental models from prior study \cite{xu2024mental}
        \item[$\Delta$] Accuracy Gain/Loss in Our Reasoning Model \cite{xu2024mental}
    \end{tablenotes}
    \end{threeparttable}
\end{table*}

\subsection{Performance Across Metrics}
Table~\ref{tab:classification_results} details additional metrics such as balanced accuracy, F1-score, and MCC. Notably:

\begin{itemize}
    \item \textbf{Few-Shot CoT consistently improves multi-class classification}, achieving the best F1-score for CSSRS (\textbf{0.438}) and DepSeverity (\textbf{0.412}).
    \item \textbf{CoT and SC-CoT enhance binary classification}, with the highest MCC (\textbf{0.623}) in Dreaddit.
    \item \textbf{ToT performs inconsistently}, excelling in precision (CSSRS, \textbf{0.579}) but lagging in accuracy.
\end{itemize}

\begin{figure}[htbp]
    \centering
    \includegraphics[width=0.9\linewidth]{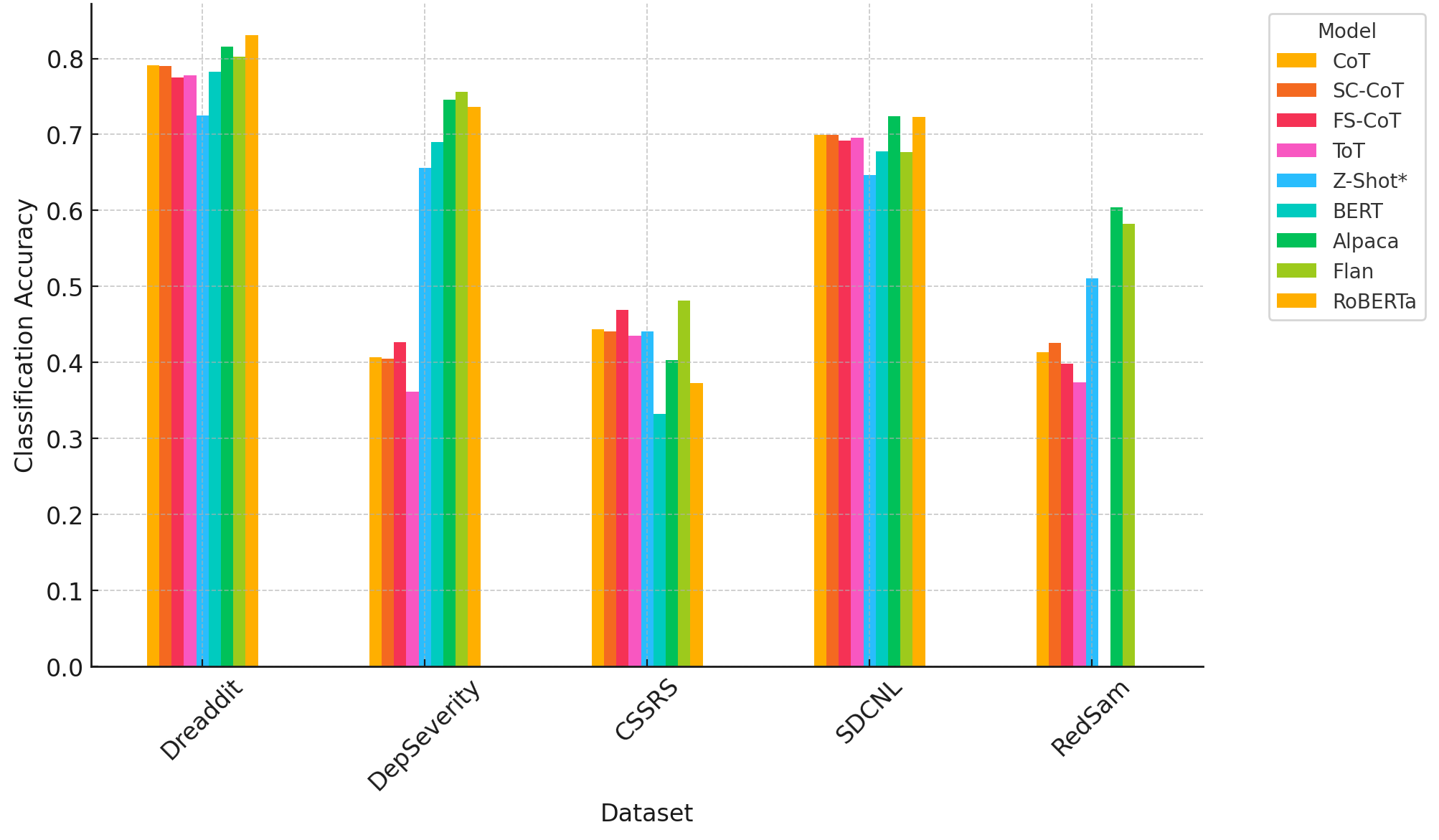}
    \caption{Macro-averaged accuracy distributions for all models across datasets, demonstrating: (1) 8-12\% gains from reasoning strategies in structured datasets (SDCNL/Dreaddit), (2) Zero-shot superiority in RedSam's imbalanced classification.}
    \label{fig:clfacc}
\end{figure}

Figure~\ref{fig:clfacc} illustrates classification accuracy comparisons across models, providing a clearer performance breakdown.

\begin{table*}[htbp]
\centering
\renewcommand{\arraystretch}{1.1}
\sisetup{
    table-format=1.3,
    table-number-alignment=center,
    detect-weight=true
}
\caption{Classification Metrics Across Datasets and Reasoning Strategies (Best Results in Bold)}
\label{tab:classification_results}
\begin{threeparttable}
\begin{tabular}{@{}l l *{10}{S}@{}}
\toprule
\multirow{2}{*}{\textbf{Dataset}} & \multirow{2}{*}{\textbf{Strategy}} & \textbf{Accuracy} & \textbf{Bal. Acc.} & \textbf{Precision} & \textbf{Recall} & \textbf{F1} & \textbf{MCC} & \textbf{MAE} & \textbf{QWK} & \textbf{ROC} & \textbf{PR} \\
& & & & \textbf{(Macro)} & \textbf{(Macro)} & \textbf{(Macro)} & & & & \textbf{AUC} & \textbf{AUC} \\
\midrule
\multirow{4}{*}{CSSRS} 
 & CoT & 0.474 & 0.445 & 0.519 & 0.445 & 0.404 & 0.316 & 0.924 & 0.407 & {--} & {--} \\
 & SC-CoT & 0.466 & 0.441 & 0.522 & 0.441 & 0.400 & 0.304 & 0.928 & 0.413 & {--} & {--} \\
 & Few-Shot & \textbf{0.492} & \textbf{0.469} & 0.540 & \textbf{0.469} & \textbf{0.438} & \textbf{0.352} & 0.908 & \textbf{0.433} & {--} & {--} \\
 & ToT & 0.476 & 0.436 & \textbf{0.579} & 0.436 & 0.392 & 0.306 & \textbf{0.888} & 0.422 & {--} & {--} \\
\midrule
\multirow{4}{*}{DepSeverity}
 & CoT & 0.527 & 0.407 & 0.378 & 0.407 & 0.355 & 0.261 & 0.791 & 0.386 & {--} & {--} \\
 & SC-CoT & 0.527 & 0.406 & 0.378 & 0.406 & 0.355 & 0.261 & 0.787 & 0.391 & {--} & {--} \\
 & Few-Shot & \textbf{0.642} & \textbf{0.428} & \textbf{0.430} & \textbf{0.428} & \textbf{0.412} & \textbf{0.319} & \textbf{0.528} & \textbf{0.494} & {--} & {--} \\
 & ToT & 0.411 & 0.362 & 0.408 & 0.362 & 0.315 & 0.194 & 0.815 & 0.371 & {--} & {--} \\

\midrule
\multirow{4}{*}{Dreaddit}
 & CoT & \textbf{0.799} & \textbf{0.791} & \textbf{0.834} & \textbf{0.791} & \textbf{0.790} & \textbf{0.623} & {--} & {--} & \textbf{0.791} & \textbf{0.727} \\
 & SC-CoT & 0.798 & \textbf{0.791} & 0.832 & \textbf{0.791} & \textbf{0.790} & 0.622 & {--} & {--} & \textbf{0.791} & \textbf{0.727} \\
 & Few-Shot & 0.784 & 0.775 & 0.827 & 0.775 & 0.772 & 0.600 & {--} & {--} & 0.775 & 0.711 \\
 & ToT & 0.787 & 0.779 & 0.829 & 0.779 & 0.776 & 0.606 & {--} & {--} & 0.779 & 0.715 \\

\midrule
\multirow{4}{*}{RedSam}
 & CoT & 0.406 & 0.413 & 0.425 & 0.413 & 0.332 & 0.068 & 0.671 & 0.191 & {--} & {--} \\
 & SC-CoT & \textbf{0.409} & \textbf{0.427} & \textbf{0.427} & \textbf{0.427} & \textbf{0.335} & \textbf{0.071} & \textbf{0.665} & \textbf{0.198} & {--} & {--} \\
 & Few-Shot & 0.366 & 0.399 & 0.420 & 0.399 & 0.321 & 0.039 & 0.706 & 0.185 & {--} & {--} \\
 & ToT & 0.359 & 0.375 & \textbf{0.427} & 0.375 & 0.296 & 0.002 & 0.713 & 0.164 & {--} & {--} \\

\midrule
\multirow{4}{*}{SDCNL}
 & CoT & \textbf{0.704} & \textbf{0.700} & 0.703 & \textbf{0.700} & \textbf{0.700} & \textbf{0.402} & {--} & {--} & \textbf{0.700} & \textbf{0.671} \\
 & SC-CoT & \textbf{0.704} & 0.699 & 0.703 & 0.699 & \textbf{0.700} & \textbf{0.402} & {--} & {--} & 0.699 & 0.670 \\
 & Few-Shot & 0.701 & 0.692 & \textbf{0.705} & 0.692 & 0.692 & 0.397 & {--} & {--} & 0.692 & 0.662 \\
 & ToT & 0.703 & 0.697 & 0.703 & 0.697 & 0.698 & 0.400 & {--} & {--} & 0.697 & 0.667 \\
\bottomrule
\end{tabular}

\begin{tablenotes}
\item[a] Bold values indicate best performance per metric within each dataset.
\item[b] MAE (lower is better) bolded for minimum values; other metrics (higher is better).
\item[c] All values rounded to 3 decimal places; "--" indicates unavailable metric.
\item[d] Abbreviations: Bal. Acc. = Balanced Accuracy, MCC = Matthews Correlation Coefficient, QWK = Quadratic Weighted Kappa.
\end{tablenotes}
\end{threeparttable}
\end{table*}

\subsection{Comparison with Prior Models}
Compared to transformer-based baselines:

\begin{itemize}
    \item \textit{RoBERTa} and \textit{FLAN-T5} maintain strong performance, particularly on Dreaddit and DepSeverity.
    \item \textbf{Reasoning strategies outperform \textit{BERT}} in CSSRS and SDCNL but fall short in DepSeverity.
    \item \textbf{Zero-shot prompting remains competitive} in RedSam, suggesting dataset-specific constraints.
\end{itemize}

\begin{figure}[htbp]
    \centering
    \includegraphics[width=0.9\linewidth]{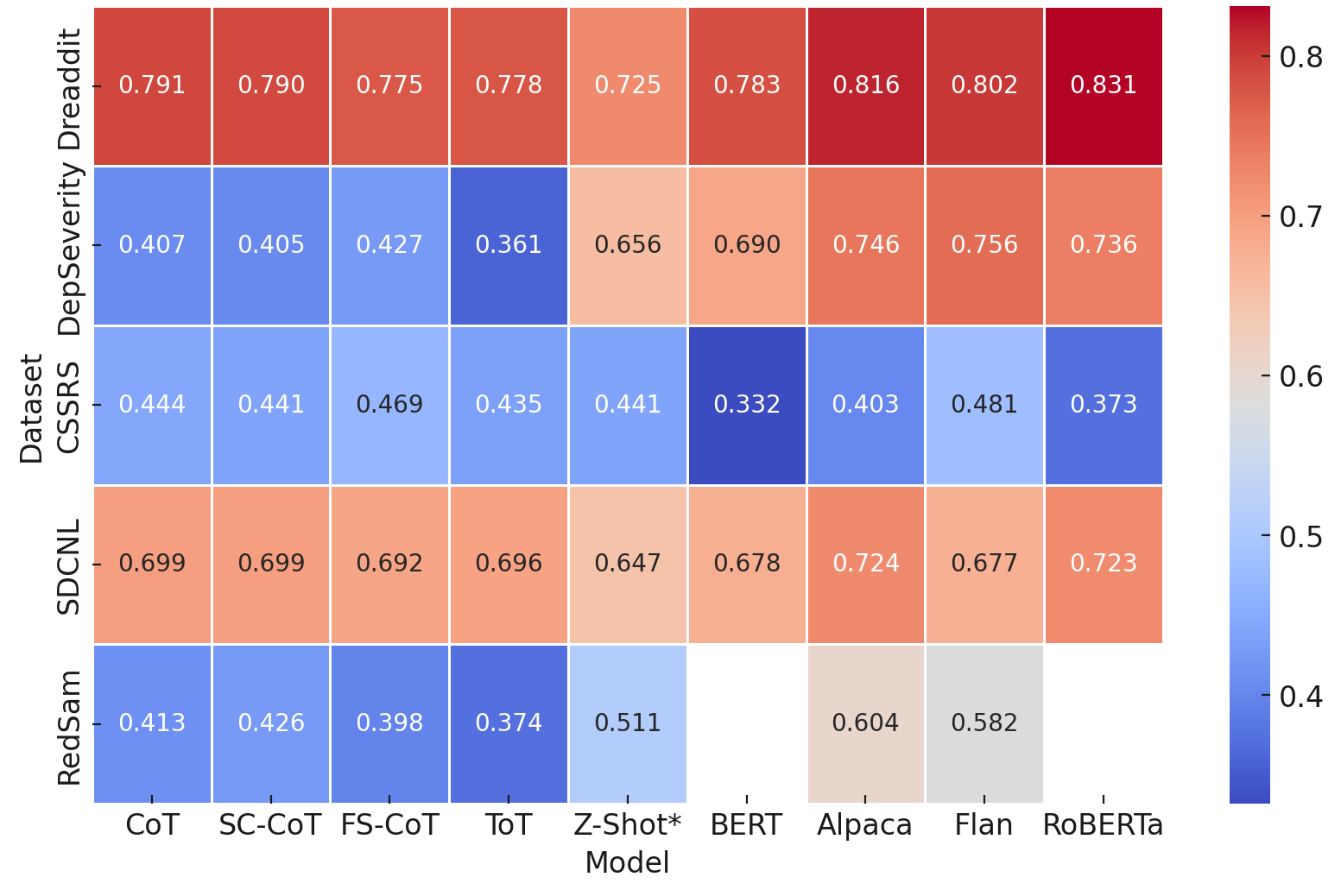}
    \caption{Accuracy heatmap comparing reasoning strategies (columns) against transformer baselines (rows) across five mental health datasets, with darker shading indicating higher performance. Highlights CoT's strong binary classification (Dreaddit/SDCNL) vs FS-CoT's multi-class advantages (CSSRS/DepSeverity).}
    \label{fig:heatmap}
\end{figure}

Figure~\ref{fig:heatmap} provides a heatmap of classification accuracy, emphasizing the variability in model effectiveness across datasets.

\subsection{Error Analysis and Failure Cases}
Despite improvements from reasoning-based techniques, several \textbf{failure patterns} emerged across datasets, revealing limitations of structured reasoning approaches.

\subsubsection{Challenges in Multi-Class Classification}
The \textbf{CSSRS} and \textbf{DepSeverity} datasets posed greater challenges than binary tasks (e.g., Dreaddit, SDCNL) due to their fine-grained labels.

\begin{itemize}
    \item \textbf{DepSeverity}: Few-Shot CoT outperformed other reasoning methods but still lagged behind zero-shot and fine-tuned transformers. A key failure was \textbf{misclassifying moderate-severity depression} as mild or severe, indicating difficulty with \textbf{nuanced classifications}.
    \item \textbf{CSSRS}: The model struggled to differentiate closely related categories (e.g., ideation vs. indicator). Few-Shot CoT performed best, but ToT produced \textbf{inconsistent reasoning paths}, lowering accuracy.
\end{itemize}

\subsubsection{Limitations in Long-Text Processing}
Datasets like \textbf{CSSRS} and \textbf{RedSam} featured \textbf{longer user-generated posts}, where CoT-based models struggled to maintain context, leading to:

\begin{itemize}
    \item \textbf{Loss of critical details}: Key linguistic cues (e.g., indirect references to suicide ideation) were sometimes overlooked in favor of superficial features, reducing classification accuracy.
    
    \item \textbf{Incomplete reasoning chains}: Although we didn't impose any output restriction but ToT responses may have been often truncated, failing to complete logical reasoning before classification.
\end{itemize}

\subsection{Summary}
While structured reasoning \textbf{enhanced classification robustness and interpretability}, challenges remain in \textbf{multi-class tasks and long-text contexts}, where nuanced reasoning and sustained context tracking are critical.

\section{Discussion and Conclusion}
\label{sec:conclusion}

\subsection{Key Findings and Performance Trends}
Our results demonstrate that structured reasoning techniques improve mental health text classification in specific scenarios, particularly for multi-class classification tasks. Among the tested prompting strategies, \textbf{Few-Shot CoT} consistently achieved higher accuracy in \textbf{CSSRS} and \textbf{DepSeverity}. In contrast, \textbf{CoT and SC-CoT} provided stable improvements for binary classification tasks such as \textbf{Dreaddit} and \textbf{SDCNL}. However, reasoning-based methods did not consistently outperform traditional transformer models, as seen in \textbf{RedSam}, where zero-shot classification remained superior.

\subsection{Comparison with Prior Work}
Our findings align with previous studies showing that CoT-style prompting enhances model reasoning capabilities~\cite{wei2022chain, wang2022self}. However, unlike general NLP tasks where CoT often surpasses traditional methods, mental health classification presents additional challenges. Prior transformer-based models such as \textbf{RoBERTa} and \textbf{FLAN-T5}~\cite{xu2024mental} outperformed our structured reasoning approaches in datasets such as \textbf{Dreaddit} and \textbf{DepSeverity}, suggesting that pre-trained classifiers still hold an advantage when domain-specific fine-tuning is involved.

\subsection{Limitations and Challenges}
While structured reasoning improves interpretability, our study highlights several challenges:

\begin{itemize}
    \item \textbf{Dataset Imbalance:} Many benchmark datasets exhibit class imbalance, particularly in \textbf{DepSeverity} and \textbf{RedSam}, where minority classes are underrepresented. This likely affected the performance of reasoning-based models, which rely heavily on prior knowledge rather than fine-tuning.
    
    \item \textbf{Multi-Class Complexity:} Few-Shot CoT performed well in \textbf{CSSRS} but struggled in \textbf{DepSeverity}. The fine-grained labels in \textbf{DepSeverity} (minimal, mild, moderate, severe) require nuanced interpretation, which reasoning-based models may not fully capture without fine-tuning.

    \item \textbf{Long vs. Short Texts:} CoT-style models excel in short, structured reasoning tasks but face limitations with lengthy social media posts, as seen in \textbf{CSSRS}, where traditional transformers outperformed reasoning-based methods.

    \item \textbf{Prompt Sensitivity:} Unlike fine-tuned models, prompting-based approaches are sensitive to minor changes in prompt phrasing.
\end{itemize}

\subsection{Implications and Future Directions}
Our study highlights the potential of structured reasoning techniques but also suggests areas for further research:

\begin{itemize}
    \item \textbf{Hybrid Models:} Combining CoT prompting with fine-tuned transformer models could leverage the strengths of structured reasoning and domain adaptation.
    \item \textbf{Automated Prompt Optimization:} Future work can explore reinforcement learning-based prompt tuning to optimize CoT effectiveness across different datasets.
    \item \textbf{Larger-Scale Reasoning Models:} Using more powerful reasoning-focused LLMs, such as OpenAI \textit{O1} or \textit{DeepSeek}, may yield further improvements over smaller models like \textit{o3-mini}.
    \item \textbf{Evaluating Reasoning Quality:} Future research could assess the coherence, logical consistency, and depth of reasoning in LLM-generated responses using a structured rubric or a numerical scale (e.g., 1–5). This could help quantify the effectiveness of different CoT prompting strategies.
\end{itemize}

\subsection{Conclusion}
While structured reasoning improves mental health classification in specific contexts, it does not universally outperform fine-tuned transformer models. Our findings suggest that \textbf{Few-Shot CoT is beneficial for multi-class classification}, while \textbf{CoT and SC-CoT are better suited for binary classification tasks}. Future research should explore hybrid approaches integrating structured reasoning with fine-tuning techniques to enhance interpretability and accuracy.

\bibliographystyle{IEEEtran}
\nocite{*}
\bibliography{ref}

@article{patil2025advancing,
  title={Advancing Reasoning in Large Language Models: Promising Methods and Approaches},
  author={Patil, Avinash},
  journal={arXiv preprint arXiv:2502.03671},
  year={2025}
}

@article{xu2024mental,
  title={Mental-llm: Leveraging large language models for mental health prediction via online text data},
  author={Xu, Xuhai and Yao, Bingsheng and Dong, Yuanzhe and Gabriel, Saadia and Yu, Hong and Hendler, James and Ghassemi, Marzyeh and Dey, Anind K and Wang, Dakuo},
  journal={Proceedings of the ACM on Interactive, Mobile, Wearable and Ubiquitous Technologies},
  volume={8},
  number={1},
  pages={1--32},
  year={2024},
  publisher={ACM New York, NY, USA}
}

@article{wei2022chain,
  title={Chain of thought prompting elicits reasoning in large language models},
  author={Wei, Jason and Wang, Xuezhi and Schuurmans, Dale and others},
  journal={Advances in Neural Information Processing Systems},
  volume={35},
  pages={24824--24837},
  year={2022}
}

@article{wang2022self,
  title={Self-consistency improves chain of thought reasoning in language models},
  author={Wang, Xuezhi and Wei, Jason and Schuurmans, Dale and Le, Quoc and Chi, Ed and Narang, Sharan and Chowdhery, Aakanksha and Zhou, Denny},
  journal={arXiv preprint arXiv:2203.11171},
  year={2022}
}

@article{yao2023tree,
  title={Tree of thoughts: Deliberate problem solving with large language models},
  author={Yao, Shunyu and Yu, Dian and Zhao, Jeffrey and Shafran, Izhak and Griffiths, Tom and Cao, Yuan and Narasimhan, Karthik},
  journal={Advances in neural information processing systems},
  volume={36},
  pages={11809--11822},
  year={2023}
}

@article{brown2020language,
  title={Language models are few-shot learners},
  author={Brown, Tom and Mann, Benjamin and Ryder, Nick and Subbiah, Melanie and Kaplan, Jared D and Dhariwal, Prafulla and Neelakantan, Arvind and Shyam, Pranav and Sastry, Girish and Askell, Amanda and others},
  journal={Advances in neural information processing systems},
  volume={33},
  pages={1877--1901},
  year={2020}
}

@article{wongkoblap2017researching,
  title={Researching mental health disorders in the era of social media: systematic review},
  author={Wongkoblap, Akkapon and Vadillo, Miguel A and Curcin, Vasa},
  journal={Journal of medical Internet research},
  volume={19},
  number={6},
  pages={e228},
  year={2017},
  publisher={JMIR Publications Toronto, Canada}
}

@inproceedings{yao2023react,
  title={React: Synergizing reasoning and acting in language models},
  author={Yao, Shunyu and Zhao, Jeffrey and Yu, Dian and Du, Nan and Shafran, Izhak and Narasimhan, Karthik and Cao, Yuan},
  booktitle={International Conference on Learning Representations (ICLR)},
  year={2023}
}

@article{mcbain2025competency,
  title={Competency of Large Language Models in Evaluating Appropriate Responses to Suicidal Ideation: Comparative Study},
  author={McBain, Ryan K and Cantor, Jonathan H and Zhang, Li Ang and Baker, Olesya and Zhang, Fang and Halbisen, Alyssa and Kofner, Aaron and Breslau, Joshua and Stein, Bradley and Mehrotra, Ateev and others},
  journal={Journal of Medical Internet Research},
  volume={27},
  pages={e67891},
  year={2025},
  publisher={JMIR Publications Toronto, Canada}
}

@article{turcan2019dreaddit,
  title={Dreaddit: A reddit dataset for stress analysis in social media},
  author={Turcan, Elsbeth and McKeown, Kathleen},
  journal={arXiv preprint arXiv:1911.00133},
  year={2019}
}

@inproceedings{naseem2022early,
  title={Early identification of depression severity levels on reddit using ordinal classification},
  author={Naseem, Usman and Dunn, Adam G and Kim, Jinman and Khushi, Matloob},
  booktitle={Proceedings of the ACM web conference 2022},
  pages={2563--2572},
  year={2022}
}

@inproceedings{sampath2022data,
  title={Data set creation and empirical analysis for detecting signs of depression from social media postings},
  author={Sampath, Kayalvizhi and Durairaj, Thenmozhi},
  booktitle={International Conference on Computational Intelligence in Data Science},
  pages={136--151},
  year={2022},
  organization={Springer}
}

@inproceedings{haque2021deep,
  title={Deep learning for suicide and depression identification with unsupervised label correction},
  author={Haque, Ayaan and Reddi, Viraaj and Giallanza, Tyler},
  booktitle={Artificial Neural Networks and Machine Learning--ICANN 2021: 30th International Conference on Artificial Neural Networks, Bratislava, Slovakia, September 14--17, 2021, Proceedings, Part V 30},
  pages={436--447},
  year={2021},
  organization={Springer}
}

@inproceedings{gaur2019knowledge,
  title={Knowledge-aware assessment of severity of suicide risk for early intervention},
  author={Gaur, Manas and Alambo, Amanuel and Sain, Joy Prakash and Kursuncu, Ugur and Thirunarayan, Krishnaprasad and Kavuluru, Ramakanth and Sheth, Amit and Welton, Randy and Pathak, Jyotishman},
  booktitle={The world wide web conference},
  pages={514--525},
  year={2019}
}

@article{posner2008columbia,
  title={Columbia-suicide severity rating scale (C-SSRS)},
  author={Posner, Kent and Brent, D and Lucas, C and Gould, M and Stanley, B and Brown, G and Fisher, P and Zelazny, J and Burke, A and Oquendo, MJNY and others},
  journal={New York, NY: Columbia University Medical Center},
  volume={10},
  pages={2008},
  year={2008}
}

@article{regier2013dsm,
  title={The DSM-5: Classification and criteria changes},
  author={Regier, Darrel A and Kuhl, Emily A and Kupfer, David J},
  journal={World psychiatry},
  volume={12},
  number={2},
  pages={92--98},
  year={2013},
  publisher={Wiley Online Library}
}

@article{world2022mental,
  title={World mental health report: Transforming mental health for all},
  author={World Health Organization},
  year={2022},
  publisher={World Health Organization}
}

@inproceedings{de2014mental,
  title={Mental health discourse on reddit: Self-disclosure, social support, and anonymity},
  author={De Choudhury, Munmun and De, Sushovan},
  booktitle={Proceedings of the international AAAI conference on web and social media},
  volume={8},
  number={1},
  pages={71--80},
  year={2014}
}

@article{calvo2017natural,
  title={Natural language processing in mental health applications using non-clinical texts},
  author={Calvo, Rafael A and Milne, David N and Hussain, M Sazzad and Christensen, Helen},
  journal={Natural Language Engineering},
  volume={23},
  number={5},
  pages={649--685},
  year={2017},
  publisher={Cambridge University Press}
}

@inproceedings{devlin2019bert,
  title={Bert: Pre-training of deep bidirectional transformers for language understanding},
  author={Devlin, Jacob and Chang, Ming-Wei and Lee, Kenton and Toutanova, Kristina},
  booktitle={Proceedings of the 2019 conference of the North American chapter of the association for computational linguistics: human language technologies, volume 1 (long and short papers)},
  pages={4171--4186},
  year={2019}
}

@article{liu2019roberta,
  title={Roberta: A robustly optimized bert pretraining approach},
  author={Liu, Yinhan and Ott, Myle and Goyal, Naman and Du, Jingfei and Joshi, Mandar and Chen, Danqi and Levy, Omer and Lewis, Mike and Zettlemoyer, Luke and Stoyanov, Veselin},
  journal={arXiv preprint arXiv:1907.11692},
  year={2019}
}

@article{perou2013mental,
  title={Mental health surveillance among children—United States, 2005--2011},
  author={Perou, Ruth and Bitsko, Rebecca H and Blumberg, Stephen J and Pastor, Patricia and Ghandour, Reem M and Gfroerer, Joseph C and Hedden, Sarra L and Crosby, Alex E and Visser, Susanna N and Schieve, Laura A and others},
  year = {2013}
}

@misc{o3-mini,
  author       = {OpenAI},
  title        = {{o3-mini}},
  howpublished = {\url{https://openai.com/index/openai-o3-mini/}},
  note         = {Accessed: Mar. 12, 2025}
}

@article{jadon2025enhancing,
  title={Enhancing Domain-Specific Retrieval-Augmented Generation: Synthetic Data Generation and Evaluation using Reasoning Models},
  author={Jadon, Aryan and Patil, Avinash and Kumar, Shashank},
  journal={arXiv preprint arXiv:2502.15854},
  year={2025}
}

@article{jadon2025ethical,
  title={Ethical AI development: Mitigating bias in generative models},
  author={Jadon, Aryan},
  journal={Interplay of Artificial General Intelligence with Quantum Computing: Towards Sustainability},
  pages={123--136},
  year={2025},
  publisher={Springer}
}
\vspace{12pt}
\end{document}